\pdfoutput=1

\documentclass[11pt]{article}
\usepackage{acl}
\usepackage{times}
\usepackage{latexsym}

\newcommand{\modelname}{\textsc{HydraSum}}

\usepackage{hyperref}
\usepackage{url}
\usepackage{graphicx}
\usepackage{array}
\usepackage{multirow}
\usepackage{url}
\usepackage{amsmath}
\usepackage{todonotes}
\usepackage{booktabs, wrapfig}
\usepackage{enumitem}
\usepackage{amssymb}
\usepackage{bbm}
\usepackage{subfig}

\newcolumntype{P}[1]{>{\centering\arraybackslash}p{#1}}

\usepackage{color, colortbl}
\definecolor{Gray}{gray}{0.9}

\usepackage[T1]{fontenc}

\usepackage[utf8]{inputenc}

\usepackage{microtype}

\title{\modelname: Disentangling Style Features in Text Summarization \\ with Multi-Decoder Models}

\author{Tanya Goyal$^1$ \hspace{0.3cm} Nazneen Rajani$^2$ \hspace{0.3cm} Wenhao Liu$^3$ \hspace{0.3cm} Wojciech Kryściński$^4$ \\
$^1$ Department of Computer Science, The University of Texas at Austin \\ $^2$ Hugging Face \hspace{0.3cm} $^3$ Faire \hspace{0.3cm} $^4$ Salesforce Research \\
{\tt tanyagoyal@utexas.edu}}

\begin{document}
\maketitle
\begin{abstract}
Summarization systems make numerous ``decisions'' about summary properties during inference, e.g. degree of copying, specificity and length of outputs, etc. However, these are implicitly encoded within model parameters and specific styles cannot be enforced. To address this, we introduce \modelname, a new summarization architecture that extends the single decoder framework of current models to a mixture-of-experts version with multiple decoders. We show that \modelname's multiple decoders automatically learn contrasting summary styles when trained under the standard training objective without any extra supervision. Through experiments on three summarization datasets (\textsc{Cnn}, \textsc{Newsroom} and \textsc{XSum}), we show that \modelname~provides a simple mechanism to obtain stylistically-diverse summaries  by sampling from either individual decoders or their mixtures, outperforming baseline models. Finally, we demonstrate that a small modification to the gating strategy during training can  enforce an even stricter style partitioning, e.g. high- vs low-abstractiveness or high- vs low-specificity, allowing users to sample from a larger area in the generation space and vary summary styles along multiple dimensions.\footnote{Code and model checkpoints are shared at \url{https://github.com/salesforce/hydra-sum}.}
\end{abstract}

\section{Introduction}
Abstractive summarization \citep{rush2015neural,see2017get} involves a combination of generation decisions, such as what content to directly copy from the input and what content to paraphrase, the level of specificity vs generality, length, readability, etc. of generated summaries. Current summarization systems \citep{lewis2020bart, zhang2020pegasus} implicitly encode these decisions in their parameters, but provide no mechanism for end users to specify their stylistic preferences. Commonly used decoding methods such as beam search, top-\textit{k} decoding \citep{fan2018hierarchical} or diverse decoding \citep{vijayakumar2018diverse} tend to generate stylistically similar outputs, and cannot be queried for multiple diverse summaries without sacrificing quality. 
Prior work in style transfer \cite{hu2017toward, krishna2020reformulating} target styles that are not relevant to summarization (e.g. sentiment, Shakespearean language, etc.) and use explicit interventions to enforce style. Instead, we ask: \textbf{what style combinations naturally occur in abstractive summarization datasets and can models automatically disentangle them?} 

\begin{figure*}[t]
\centering
    \includegraphics[trim=8mm 120mm 0mm 20mm,scale=0.24, clip]{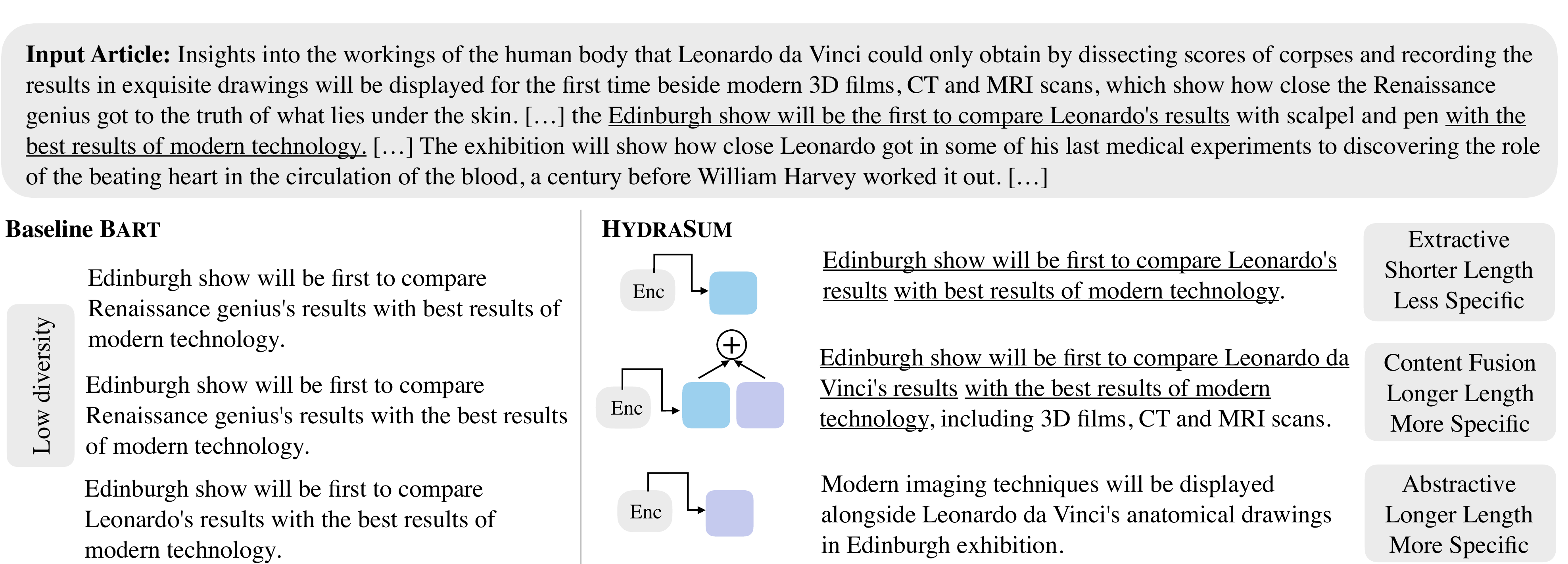}
    \caption{Examples of generated summaries for a \textsc{Newsroom} article using both \textsc{Bart} and a 2-decoder \modelname~model. Longer copied sequences (denoting extractive behavior) are underlined. For \modelname, summaries from different mixtures of decoders differ in degree of abstractiveness, specificity, and length.}
    \label{fig:example-intro}
\end{figure*} 

In this paper, we propose \modelname~- a new summarization architecture that disentangles the different stylistic decisions made by abstractive summarization models from the models weights into an explicit model component. Our model contains a single transformer-based encoder to encode the input document and a mixture-of-experts with multiple decoders for summary generation. At each time step of the generation phase, the next token's probability distribution is computed by combining the output probabilities of all individual decoders. This allows our model to distribute the diverse stylistic and lexical features encountered in the training data, even those within the same reference summary, across the parameters  of separate decoders. During inference, we leverage the modularity in the decoder framework to sample from these individual decoders, each of which generates stylistically-distinct summaries.

As a toy example, consider a 2-decoder scenario in which one decoder learns to only copy phrases or words from the input document, while the second decoder only learns paraphrasing and syntactic transformations. While individual decoders cannot cover the range of stylistic variations in the dataset, a weighted combination or mixture of the two decoders can be used to model the summarization dataset. In practice, we found that this partitioning of summarization ``skills'' between decoders is much fuzzier, and occurs along multiple dimensions such as degree of abstractiveness (copying), readability, specificity and length. Figure \ref{fig:example-intro} shows examples of  summaries generated by the baseline model and a 2-decoder version of \modelname. We see that \modelname~produces a more stylistically distinct set of summaries by varying the degree of abstractiveness and summary length, or including additional details such as \textit{3D films, CT and MRI scans} to vary specificity. On the other hand, baseline \textsc{Bart} exhibits low diversity and largely generates extractive summaries \cite{see2017get, goyal-durrett-2021-annotating}.

Our contributions in this paper are: (1) We show that our proposed \modelname~model automatically assigns distinct summary ``\emph{skills}'' to different decoders during training, for both 2- and 3-decoder versions across three summarization datasets (Section~\ref{sec:experiments-unguided}). (2) We show that this property can be operationalized to obtain multiple summaries exhibiting better stylistic diversity and Top-K quality compared to baseline models (Section~\ref{sec:diversity}). (3) Finally, we demonstrate that a simple data pre-processing and gating strategy during training can be used to explicitly dictate \textit{which} feature is partitioned across different decoders. Not only does this allow us to enforce a greater style difference between decoders compared to prompt-based baselines, it also provides a mechanism for multi-style variation in summary generation (Section~\ref{sec:experiments-guided}).

\begin{figure}[t]
\centering
    \includegraphics[trim=260mm 54mm 0mm 40mm,scale=0.28, clip]{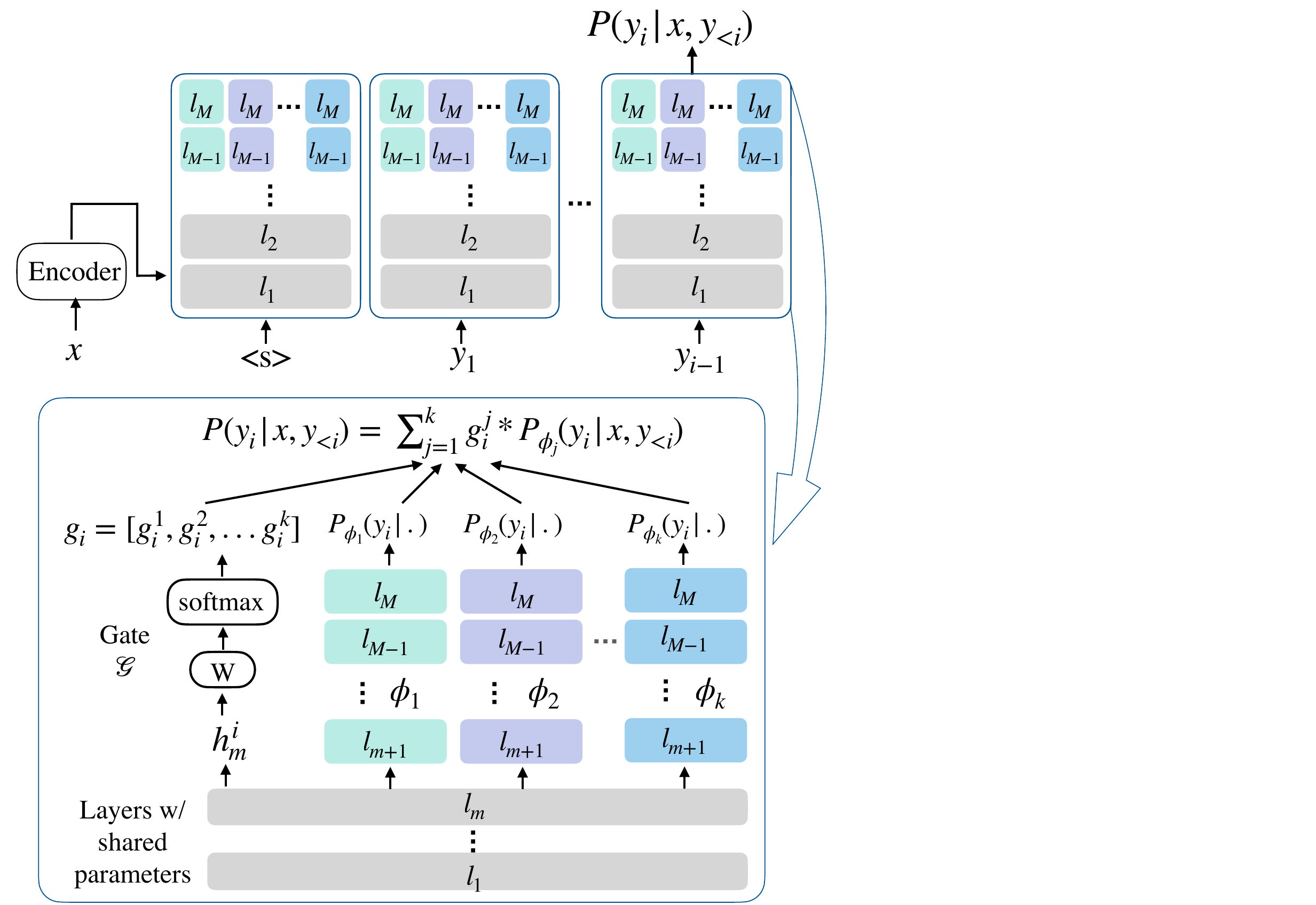}
    \caption{Our proposed \modelname~architecture. The decoder network of standard models is modified to incorporate multiple decoders. The lower layers of these decoders have shared parameters and a gating mechanism is used to combine their output probabilities in a mixture-of-experts formulation.} 
    \label{fig:model-arch}
\end{figure}

\section{Methodology}
Current state-of-the-art summarization models (e.g. \textsc{Bart}, \textsc{Pegasus}) use transformer-based encoder-decoder architectures. Similarly to those models, \modelname~consists of an encoder network that accepts the document $x$ as input. The decoder network, however, is modified to incorporate $k(>1)$ decoders,  $\phi_1, \phi_2, ... \phi_k$, as depicted in Figure  \ref{fig:model-arch}. At time step $i$, each decoder outputs a probability distribution $P_{\phi_k}(y_i|x, y_{<i})$ over the vocabulary, corresponding to the next-token probabilities. The final output probability $P(y_{i}|x, y_{<i})$ is computed as a mixture of these $k$ probability distributions, with the mixing coefficients predicted by a gating mechanism $\mathcal{G}$.

\paragraph{Multi-Decoder Architecture} Let $M$ be the total number of decoder blocks in a single decoder: e.g. $M=12$ for \textsc{Bart-Large}. In \modelname, the parameters of the $m (< M)$ bottom layers are shared between the $k$ decoders. This reduces the number of extra parameters introduced into the model architecture. The top $M-m$ layers of the different decoders are independently trained. The right block of Figure \ref{fig:model-arch} shows a detailed view of the multi-decoder architecture at a single time step $i$.

\paragraph{Gating Mechanism} A gating mechanism $\mathcal{G}$ is used to combine the output distributions of the $k$ decoders. Let $h_i^m$ be the hidden state output of the $m^{th}$ decoder layer at time step $i$, i.e. the output of the last shared layer. We use this hidden state representation to obtain the coefficients for our mixture of experts. The representation $h_i^m$ is fed into a feed forward layer $W$ (size $= (|h_i^m|, k)$), followed by a softmax layer. This outputs a probability distribution $g_i$ which is used to compute the overall next-token output probability as follows:
$P(y_i|x, y_{<i}) = \sum\nolimits_{j=1:k}{g_i^j * P_{ \phi_j}(y_i|x, y_{<i})}$. Here, $g_i^j$ is the probability of selecting the $j^{th}$ decoder at time step $i$.

\paragraph{Training} Similar to standard summarization models, the \modelname~architecture is trained to minimize the cross entropy loss of the reference summaries, conditioned on the input document: $loss = -\sum\nolimits_i\log P(y_i|x, y_{<i})$. The model implicitly decides the contribution of each decoder to the final output probability, i.e. $g_i^j$ for decoder $j$ at time step $i$, using the gating mechanism $\mathcal{G}$ from above. 

\begin{figure}[t]
\centering
    \includegraphics[trim=32mm 253mm 300mm 25mm,scale=0.24, clip]{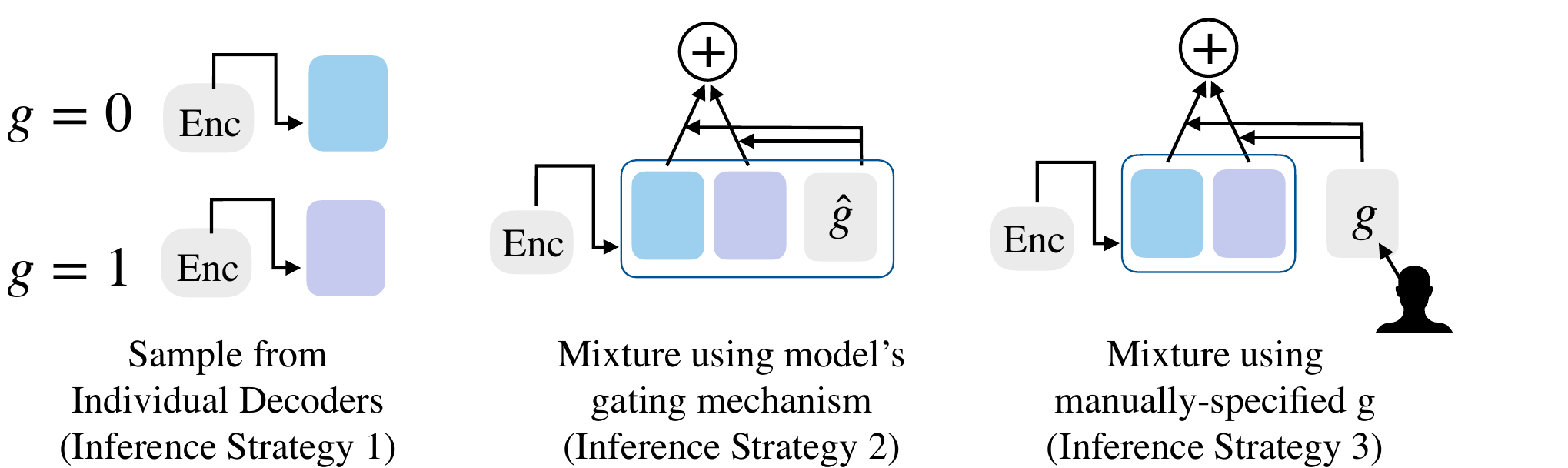}
    \caption{\modelname's inference options.} 
    \label{fig:inference-strategies}
\end{figure} 

\subsection{Inference} 
\label{sec:inference-settings}
\modelname~provides several options of output distributions which differ in how the mixture weights are obtained (see Figure \ref{fig:inference-strategies}). During inference, we can sample from these different options, or \textbf{inference strategies}, to generate summaries:
\begin{enumerate}[leftmargin=*]
    \item \textbf{Individual Decoders}: To generate summaries using only the $j^{th}$ decoder, the output of the gating mechanism is overridden with $[0,0...,1,...,0]$ where $g^j = 1$ and $g^{i \neq j} = 0$ for all time steps. 
    \item \textbf{Mixture using $\mathbf{\mathcal{G}}$}: The mixture weights are decided by the model, i.e. $g_i^j = (W^T h_i^m)_j$ for decoder $\phi_j$ at time step $i$.
    \item \textbf{Mixture with manually-specified $\mathbf{g}$}: Consider a 2-decoder \modelname~model, where decoder 0 learns abstractive and decoder 1 learns extractive features. The degree of abstraction can be varied by specifying the contribution of individual decoders through gate coefficients $[1-g, g]$. Effectively, this modifies the output probability to: $P(y_i|\cdot) = (1- g) * P_{\phi_0}(y_i|\cdot) + g * P_{\phi_1}(y_i|\cdot)$.
\end{enumerate}

\section{Experiments}
\label{sec:experiments}
We conduct experiments on three news summarization datasets: \textsc{Cnn} \citep{hermann2015teaching, nallapati2016abstractive}, \textsc{Newsroom}\footnote{We run experiments on the \textit{mixed} subset of \textsc{Newsroom} to limit data size. We found that this subset was less noisy and more diverse than the \textit{abstractive} and \textit{extractive} subsets.} \citep{grusky2018newsroom} and \textsc{XSum} \citep{narayan2018don}. The reference summaries in these datasets exhibit a mutually-distinct stylistic properties and help evaluate \modelname's capabilities under these distinct test conditions. 

For all experiments, \textsc{Bart-Large} \citep{lewis2020bart} is used as the model initialization: in a $k$-decoder variant of \modelname, all $k$ decoders are initialized with the weights of \textsc{Bart-Large}'s decoder.  The weights of the gating mechanism $\mathcal{G}$ are randomly initialized from a normal distribution $\mathcal{N}(0, 0.02)$. We set the number of shared layers, i.e. $m$ to 8, for all experiments.\footnote{Experiments with other values of $m(=6, 10)$ are in Appendix \ref{appendix:varying-m}. Varying $m$ does not alter our conclusions.} Our model architecture is implemented using the Huggingface Library \citep{wolf2020transformers}. More training and inference details are in Appendix \ref{appendix-trainingdetails}.

We compare against the standard \textsc{Bart}-based summarization baseline. For \textsc{XSum}, we use the publicly available \textsc{Bart-Large-XSum} checkpoint. For \textsc{Cnn} and \textsc{Newsroom}, we fine-tune the \textsc{Bart-Large} checkpoint on their corresponding training datasets ourselves.\footnote{Publicly available \textsc{Bart-Large-Cnn} \citep{lewis2020bart} and \textsc{Pegasus-Newsroom} \citep{zhang2020pegasus} trained on the full \textsc{CnnDm} and \textsc{Newsroom} datasets perform poorly on the \textsc{Cnn} only and \textsc{Newsroom-Mixed} only test sets used in our work. Hence, we re-train these.} Beam decoding is used to generate summaries for all models.


\subsection{Style Partitioning}
\label{sec:experiments-unguided}
First, we investigate whether individual \modelname~decoders learn  different styles when trained using the standard training objective? If yes, which stylistic features vary across different decoders? 

\paragraph{Metrics} We measure \textit{style} along the following summarization-relevant dimensions: 
\begin{enumerate}[leftmargin=*]
    \item \textbf{Abstractiveness}: We follow \citet{grusky2018newsroom} and report two metrics, \textit{coverage} which denotes the fraction of summary words that are also present in the input, and \textit{density} which denotes the average length of copied contiguous spans in a summary. Additionally, we report the 2-gram overlap between the generated summary and the input article.
    \item \textbf{Degree of specificity} of generated summaries, quantified using the Speciteller tool \citep{li2015fast}. To align with their definition, we segment summaries into sentences and report the macro-average of the sentence-level specificity across all summaries.
    \item \textbf{Length metrics}: We report two metrics for this, \textit{absolute length} (number of words) of generated summaries, and \textit{compression ratio}, computed as the ratio of the number of words in the summary and the input article.
    \item \textbf{Readability} scores of generated summaries, measured using the Flesch readability ease test \citep{flesch1948new}.
\end{enumerate}

In addition to these style-based metrics, we report \textbf{Quality}, measured by \textsc{Rouge} \citep{rouge2004package} scores of the generated summaries with respect to the reference summaries. 


For analysis, we generate 3 summaries for each input: using individual decoders D0 and D1 (Inference Strategy 1, see Section \ref{sec:inference-settings}), and the mixture model (Mix) where the mixture weights are obtained using the gating mechanism $\mathcal{G}$ (Strategy 2). The latter corresponds to sampling from the \modelname's actual output distribution. 

\begin{table*}[t]
    \small
    \renewcommand{\tabcolsep}{1.6mm}
    \centering
    \begin{tabular}{c|r|ccc|c|cc|c|c}
    \toprule
    &  & \multicolumn{3}{c|}{Abstractiveness} & Specificity & \multicolumn{2}{c|}{Length-metrics} & Readability & Quality \\
    & & Coverage & Density & 2G Overlap & & Abs. & Comp. & FRE & \textsc{R1/R2/RL} \\
    \midrule
    \multirow{5}{*}{\textbf{\textsc{Cnn}}} & Ref & 0.85 & 3.14 & 0.43 & 0.44 & 37.33 & 0.07 & 52.51 & - \\
    & Baseline & 0.97 & 10.33 & 0.80 & 0.44 & 50.71 & 0.10 & 54.03 & 34.87/\textbf{14.88}/31.82\\ \cmidrule{2-10}
    & D0  & \cellcolor{Gray}0.93 & \cellcolor{Gray}5.69 & \cellcolor{Gray}0.64 & \cellcolor{Gray}0.48 & \cellcolor{Gray}46.07 & \cellcolor{Gray}0.09 & \cellcolor{Gray}58.00 & 34.58/13.64/31.43\\
    & D1  & \cellcolor{Gray}0.97 & \cellcolor{Gray}11.69 & \cellcolor{Gray}0.82 & \cellcolor{Gray}0.40 & \cellcolor{Gray}59.47 & \cellcolor{Gray}0.11 & \cellcolor{Gray}50.92 & 31.44/11.72/28.58\\
    & Mix  & 0.97 & 11.1 & 0.81 & 0.46 & 54.66 & 0.10 & 53.7 & \textbf{34.91/}14.36\textbf{/31.93} \\
    \midrule
    \multirow{5}{*}{\textbf{\textsc{NRoom}}} & Ref &  0.83 & 3.40 & 0.46 & 0.57 & 23.67 & 0.07 & 50.8 & - \\
    & Baseline & 0.96 & 14.34 &  0.80 & 0.63 & 34.11 & 0.10 & 48.64 & \textbf{36.38/19.54/31.20} \\ \cmidrule{2-10}
    & D0 & \cellcolor{Gray}0.90 & \cellcolor{Gray}6.15 & \cellcolor{Gray}0.59 & \cellcolor{Gray}0.65 & 33.95 & 0.10 & 49.58 & 34.64/16.59/28.94 \\
    & D1 & \cellcolor{Gray}0.96 & \cellcolor{Gray}16.45 & \cellcolor{Gray}0.84 & \cellcolor{Gray}0.58 & 34.66 & 0.10 & 49.41 & 33.73/17.27/28.90 \\
    & Mix  & 0.96 & 17.13 & 0.81 & 0.63 & 38.34 & 0.11 & 48.38 & 35.32/18.69/30.31 \\
    \midrule
    \multirow{5}{*}{\textbf{\textsc{Xsum}}}& Ref & 0.66 & 1.05 & 0.16 & 0.65 & 21.1 & 0.09 & 59.6 & -\\
    & Baseline & 0.75 & 1.61 & 0.27 & 0.56 & 19.20 & 0.09 & 66.70 & \textbf{45.14/22.27/37.25} \\ \cmidrule{2-10}
    & D0  & 0.72 & 1.37 & 0.23 & \cellcolor{Gray}0.66 & 19.72 & 0.09 & 60.45 & 42.82/19.16/34.15 \\
    & D1  & 0.72 & 1.44 & 0.23 & \cellcolor{Gray}0.53 & 19.96 & 0.09 & 62.70 & 42.33/18.56/33.98 \\
    & Mix  & 0.73 & 1.51 & 0.25 & 0.59 & 19.60 & 0.10 & 62.07 & 44.72/21.47/36.36 \\
    \bottomrule
    \end{tabular}
    \caption{Comparison of \modelname's generated summaries using individual decoders (D0 and D1) and their model-derived mixture (Mix). Results show significant differences along multiple dimensions (highlighted in gray), most notably abstractiveness and specificity for \textsc{Cnn} and \textsc{Newsroom}, and specificity for \textsc{XSum}.}
    \label{table:experiments-unguided}
\end{table*}

\subsection{Results}
\paragraph{Style differences between decoders} Differences in style between D0 and D1 are outlined in Table \ref{table:experiments-unguided}. Features for which this difference is significant, i.e. $p < 0.05$ according to the bootstrap re-sampling test, are highlighted in gray. For both \textsc{Cnn} and \textsc{Newsroom}, significant differences are observed along the abstractiveness and specificity metrics. Moreover, summaries for \textsc{Cnn} also differ along other metrics such as length and readability. The least amount of style difference is observed for \textsc{XSum} where the decoders only differ in specificity, although this difference (approx. .13) is more than the other datasets. We hypothesize that the similarity in abstractiveness levels of the \textsc{XSum} decoders is due to the low diversity along this feature in \textsc{XSum}'s training data. These results indicate that although \modelname's training encourages the two decoders to learn distinct styles, the combination of features along which they differ is heavily dependent on the datasets themselves.

\begin{figure}[t]
\centering
    \includegraphics[trim=105mm 100mm 0mm 80mm,scale=0.19, clip]{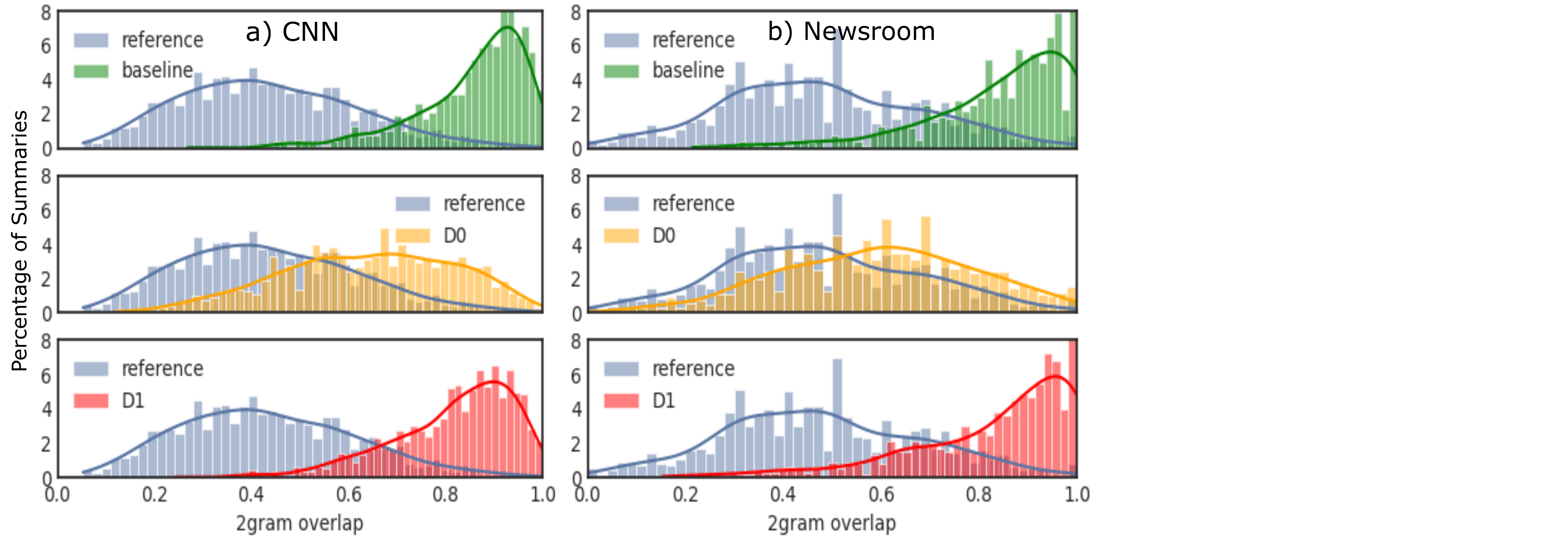} 
    \caption{Graphs plot the 2gram overlap of the baseline and \modelname~decoders. Compared to the baseline, D0 decoder samples summaries from a distribution that more closely resembles the reference distribution.}
    \label{fig:lexical-unguided}
\end{figure} 

\paragraph{Coverage over the generation space}
Interestingly, for both \textsc{Cnn} and \textsc{Newsroom}, we observe that the baseline model fails to cover the entire range of abstractive behavior seen in the reference summaries. Figure \ref{fig:lexical-unguided} demonstrates this; the top graphs plot the 2-gram overlap of the reference summaries and the baseline \textsc{Bart} summaries, showing substantial mismatch. The references are more diverse, while \textsc{Bart} summaries are highly extractive. This is a known issue with standard training \citep{see2017get, goyal2021training}; summarization models tend to overfit on the easier extractive examples and do not learn from the abstractive examples. \modelname~addresses this limitation by encouraging the two decoders to learn contrasting levels of abstractiveness. Figure \ref{fig:lexical-unguided} shows that the D0 decoders for both datasets generate abstractive summaries that more closely resembles the reference distribution. Meanwhile, D1 generates extractive summaries, collectively providing better coverage over the abstractiveness space. Later, in Section \ref{sec:experiments-guided}, we show that we can reliably vary abstractiveness between these two decoder levels using their mixture.

\paragraph{How do \modelname~decoders learn different style features?} Note that we do not introduce constraints or differ the training of the two decoders in any way; this stylistic partitioning naturally emerges. In fact, both decoders are initialized symmetrically, with \textsc{Bart-Large}. However, the randomly initialized gate $\mathcal{G}$ assigns different weight coefficients to the two decoders in the mixture, and hence their respective contributions to the output probability is different. This ensures that the gradient updates for the two decoders start to differ from the initial stages of the training itself. Eventually, as training progresses, we see that the two decoders learn very different style features characterized by differently learnt weight parameters.\footnote{We re-run these experiments with different gate initializations; style partitioning is observed consistently across runs, although the exact degree of partitioning differs slightly.}

\paragraph{Quality} The \textsc{Rouge} scores of the generated summaries using the entire \modelname~model, i.e. Mix, are comparable to the baseline \textsc{Bart} models, even outperforming the baseline for \textsc{Cnn} (see Table \ref{table:experiments-unguided}). This shows that additional decoders in \modelname~does not hurt quality. Notably, the quality of individual decoders is roughly 2 \textsc{Rouge} points lower than both the Mix strategy. This is expected; individual decoders generate summaries that exhibit ``extreme'' or contrasting behaviors along style features (shown above). Therefore, they underperform when evaluated on the entire test set containing a diverse set of styles. 

Recent work \cite{fabbri2021summeval} has shown that \textsc{Rouge} is insufficient to evaluate summary quality and recommends human evaluation. We report these results in Section~\ref{sec:human-eval}; they show that \modelname~outperforms or is on par with the baseline for all datasets.

\begin{table*}[t]
    \centering
    \small
    \begin{tabular}{c|ccc|c}
    \toprule
        Dataset & BS + Beam & BS + Top-k & BS + DBS & HS+Beam  \\ \midrule
        \textsc{Cnn} & 39.10/17.76/35.65 & 40.29/15.37/36.14 & 40.62/18.65/37.04 & \textbf{42.07/19.19/38.32}\\
        \textsc{Newsroom} & 43.00/24.73/36.98 & 43.58/22.25/36.27 & 43.59/24.72/37.27 & \textbf{45.03/25.59/38.46} \\
        \textsc{XSum} & 50.19/\textbf{25.74}/40.86 & 48.16/21.68/37.98 & 50.52/25.72/41.06 & \textbf{51.03/}25.46\textbf{/41.18} \\
    \bottomrule
    \end{tabular}
    \caption{Diversity performance (TopK \textsc{R1/R2/RL}) of the baseline \textsc{Bart} (BS) and \modelname~(HS) models.}
    \label{tab:diversity-eval-rouge}
\end{table*}

\subsection{Diversity Evaluation}
\label{sec:diversity} 
\modelname~provides a straightforward method to sample multiple summaries from its multiple decoders and their combination. Here, we compare the quality of these diverse set of summaries.

Following prior work in diversity evaluation \citep{vijayakumar2018diverse}, we report the TopK~\textsc{Rouge} metric: the maximum \textsc{Rouge} (R1/R2/RL) score over a list of K generated summaries for a given input. This gives an upper bound on the benefit that can be derived from diverse summarization by measuring the closeness of the best generated summary to the reference summary. We set K$=5$ for our experiments. For \modelname, multiple summaries are generated by varying the summary-level gating probability $g$ (Strategy 3, Section \ref{sec:inference-settings}). We set $g = \{0, .25, .5, .75, 1\}$; here, $g=0$ and $g=1$ correspond to summaries generated using D0 and D1 independently. These are compared to K summaries sampled from the baseline \textsc{Bart} model using the following decoding strategies: beam search, top-k sampling, and diverse beam search \citep{vijayakumar2018diverse}. Decoding hyperparameters for all settings are in Appendix \ref{appendix-trainingdetails}.

Table \ref{tab:diversity-eval-rouge} outlines our results. It shows that \modelname~substantially outperforms the baseline across all different decoding strategies considered. In fact, \textbf{the gain is roughly proportional to the degree of stylistic difference observed in Table~\ref{table:experiments-unguided}}; the highest gain (roughly +3 \textsc{Rouge} points) is reported for \textsc{Cnn}, followed by an improvement of +2 \textsc{Rouge} points for the \textsc{Newsroom} dataset. 

\begin{table}[t]
\renewcommand{\tabcolsep}{1.3mm}
\small
\centering
\begin{tabular}{c|c|cccc}
\toprule
\textbf{Dataset} & \textbf{Dec.} & \textbf{Rouge (R1/R2/RL)} & \textbf{2gm} & \textbf{Spec.} & \textbf{Len.}\\ \midrule
\multirow{4}[1]{*}{\textsc{Cnn}} & D0 & 32.35/10.90/29.29 & .48 & .34 & 39.9\\
& D1 & 21.63/8.48/20.18	& .82 &	.38 & 180.7  \\
& D2 & 33.86/13.23/30.87 & .72 & .55 & 56.1 \\
& Mix & \textbf{34.30/14.38/31.36} & .82 & .48 & 56.2\\ \midrule
\multirow{4}[1]{*}{\textsc{NR}} & D0 &	31.88/14.71/27.12 & .32 & .42 &	32.0 \\
& D1 & 16.05/6.94/14.39	& .36 & .49 & 171.9 \\
& D2 &	32.43/16.57/27.61 & .85	& .67 & 47.9  \\
& Mix	& \textbf{35.39/18.85/30.37} & .82 & .64 & 38.9 \\ \midrule
\multirow{4}[1]{*}{\textsc{XSum}} & D0 & 31.63/12.21/24.83 &	.36 & .60 & 44.6 \\
& D1 & 41.86/17.97/33.22 &	.22 & .54 & 20.1 \\
& D2 & 32.33/12.63/25.44 & .32	& .67 & 44.1 \\
& Mix & \textbf{44.61/20.91/36.17} & .24 & .58 & 19.5 \\
\bottomrule
\end{tabular}
\caption{Stylistic variation between generated summaries in a 3-decoder \modelname~model. Results show higher variation between individual decoders compared to the 2-decoder version.} 
\label{table:experiments-unguided-3decoders} 
\end{table}

\subsection{Effect of number of decoders}
\label{sec:3decoder}
We investigate this by extending our analysis to a 3-decoder variant of \modelname. Table \ref{table:experiments-unguided-3decoders} outlines our results. For simpler analysis, we only report 4 metrics: \textsc{Rouge}, 2-gram overlap, specificity and absolute length. 

Similar to the 2-decoder case, the 3 decoders of \modelname~learn a mutually-distinct combination of summary styles. In fact, \textbf{3-way partitioning allows the model to cover a wider range of summary styles.} For example, the 3-decoder \modelname~model partitions along the abstractiveness feature for XSum (D0 and D2 are more extractive compared to D1), while this was not achieved by the 2-decoder variant in Table~\ref{table:experiments-unguided}. Similarly, the specificity range for \textsc{Cnn} ($.34-.55$) and \textsc{Newsroom} ($.42 - .67$) is higher compared to the 2-decoder variant. Note that some decoders report very poor quality (\textsc{Rouge} scores). This is expected as these decoders exhibit extreme summary styles (e.g. very long summaries) and therefore suffer on dataset-wide evaluation. However, across all datasets, mixture-decoding outperforms individual decoders. This shows that although the performance of some individual decoders is low, their contribution to the mixture is critical. 

\begin{figure*}[t]
\centering
    \includegraphics[trim=52mm 215mm 25mm 40mm,scale=0.30, clip]{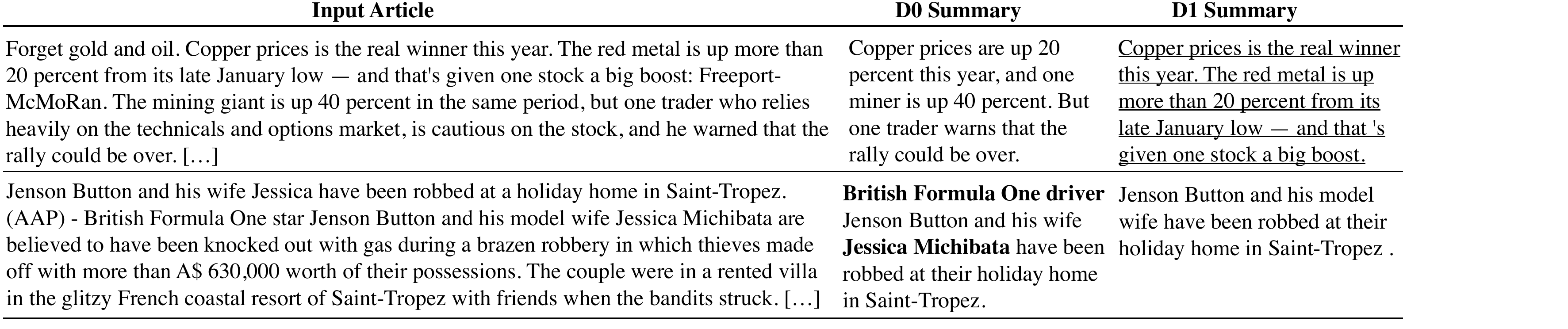} 
    \caption{Examples of \modelname~summaries from the \textsc{Newsroom} dataset. Long extractive spans are underlined, additional details that increase the specificity of summaries are in bold.} 
    \label{fig:qualitative-unguided}
\end{figure*} 

\subsection{Qualitative Evaluation} 
\label{sec:human-eval}
Figure \ref{fig:qualitative-unguided} shows examples of 
the style difference between \modelname~summaries sampled from individual decoders. In the first example, D1 generates a highly extractive summary whereas D0 generates an abstractive summary with less copying. In the second example, we observe a difference in specificity: D0 summary includes additional details like \textit{Jenson Button}'s profession and his wife's name, compared to the more general summary by D0. \modelname's architecture provides easy access to such stylistically-distinct summary sets.

\section{Extreme partitioning}
\label{sec:experiments-guided}
In Section~\ref{sec:experiments}, style partitioning was automatically driven by dataset properties. Here, we investigate whether we can explicitly dictate which specific stylistic feature differs between two decoders. Suppose our target feature (denoted by $f$) is specificity: under this scenario, we want D0 to generate low- and D1 to generate high-specificity summaries. We should also be able to generate multiple mid-specificity summaries by \textit{mixing} these two extreme decoders. In this section, we run experiments on two target features; abstractiveness (measured by 2-gram overlap) and specificity.

\paragraph{Our Method} To ensure D0 learns low-$f$ and D1 learns high-$f$, we carefully control the contribution of each training example to individual decoder's training.  Our exact methodology is: (1) First, we pre-process the training data to derive their percentile scores $p$ based on the $f$-value of reference summaries (e.g., if $f=$ abstractiveness, we use 2-gram overlap). (2) We derive $K=5$ partitions of the data based on these percentile scores. For each example, we set its oracle gate probability $g^* \in \{0, 0.25, 0.5, .75, 1\}$ to incorporate information about the percentile split it belongs to. As an example, the bottom 20 percentile of the data (low $f$) are assigned $g = 0$. (3) Next, instead of using the automatic gating mechanism $\mathcal{G}$ during training, we use the oracle label $g^*$ to derive the mixture coefficients $[1-g^*, g^*]$ and compute loss as follows: 
\begin{equation*}
\begin{aligned}
    loss = -\sum\nolimits_i\log & \big[ (1 -g^*) * P_{\phi_0}(y_i|x, y_{<i}) \\ 
    & + g^* * P_{\phi_1}(y_i|x, y_{<i})\big]
\end{aligned}
\end{equation*} 

This allows us to explicitly set the contribution of each training example to different decoders' parameter updates and ensure that D0 and D1 predominantly learn from low- and high-$f$ summaries respectively. Note that the oracles $g^*$ can be defined at the token-, sentence- or summary-level. Since specificity is defined per sentence, we derive individual oracles gates $g^*_t$ for each sentence $s_t$. For abstractiveness, we use oracle gates derived at the summary-level.

\begin{table}[t]
    \small
    \renewcommand{\tabcolsep}{0.8mm}
    \centering
    \begin{tabular}{c|c|ccc|ccc}
    \toprule
         & \textbf{Metric} $f$ & \multicolumn{3}{c|}{\textbf{Abstractiveness}} & \multicolumn{3}{c}{\textbf{Specificity}} \\ \midrule
        \textbf{Model} & & \textsc{Cnn} & \textsc{NR} & \textsc{XSum} & \textsc{Cnn} & \textsc{NR} & \textsc{XSum} \\ \midrule
    Prompt- &   $f$(``\textit{Low}'') & .68 & .62 & .21 & .44 & .53 & .52 \\
    Based & $f$(``\textit{High}'') & .83 & .84 & .24 & .53 & .76 & .69 \\ 
        \midrule
    \textsc{Hydra-} & $f$(D0) & .48 & .44 & .16 & .22  & .36 & .44 \\
    \textsc{Sum} &  $f$(D1) & .82 & .85 & .29  & .62 & .81 & .80 \\
    \bottomrule
    \end{tabular}
    \caption{Comparison between the extreme partitioning of \modelname~and the prompt-based \textsc{Bart} models.}
    \label{table:experiments-guided-small}
\end{table}

\paragraph{Baseline} We compare our model to the popular prompt-based approaches from recent controllable summarization research \cite{he2020ctrlsum}. To emulate the 2 decoder setting of \textsc{HydraSum}, we construct 2 prompts ``\textit{Low}'' and ``\textit{High}'' to indicate low- and high-$f$ respectively. We divide the training data into two subsets based on their $f$-values and train models by prepending the prompt to the reference summary. During inference, we sample 2 different summaries using these prompts and compare their $f$-difference compared to \modelname's extreme partitioning. 

\paragraph{Analysis} Table~\ref{table:experiments-guided-small} outlines our results. For each model, we report $f$(D0) and $f$(D1): the average style/feature scores for test summaries generated by D0 and D1 respectively.\footnote{Detailed results with other metrics and examples of \textit{extreme} summaries are included in Appendix \ref{appendix:guided-setting}.} 
Our results clearly show that extreme partitioning outperforms the prompt-based baselines. Moreover, it achieves better or more ``extreme'' partitioning along the target $f$ compared to \modelname~decoders in Table \ref{table:experiments-unguided}.

Figure \ref{fig:examples-specificity} shows examples of generated summaries using the extreme specificity decoders. The high specificity D1 decoder tends to include more details compared to summaries generated using D0.

\begin{table}[t]
\small
\begin{tabular}{p{3.3cm}|p{3.6cm}}
\toprule
\hfil \textbf{Low Spec. Decoder (D0)} &  \hfil \textbf{High Spec. Decoder (D1)}
\\ \midrule
Two Florida boys are being hailed as local heroes after saving children from a burning mobile home
 & \underline{Isiah Francis, 10}, and \underline{Jeremiah} \underline{Grimes, 11}, saved two babies from a burning mobile home in Florida. \\
 \rowcolor{Gray}
 French prosecutor says he is not aware of any video footage from on board the plane. & French prosecutor says he's not aware of any video footage from on board \underline{Germanwings Flight 9525}. \\
\bottomrule
\end{tabular}
\caption{Example summaries generated using low and high specificity decoders when $f=$  specificity. Extra details in more specific summaries is underlined.} 
    \label{fig:examples-specificity}
\end{table}

\begin{figure*}[h]
\centering
    \includegraphics[trim=40mm 170mm 40mm 40mm,scale=0.27, clip]{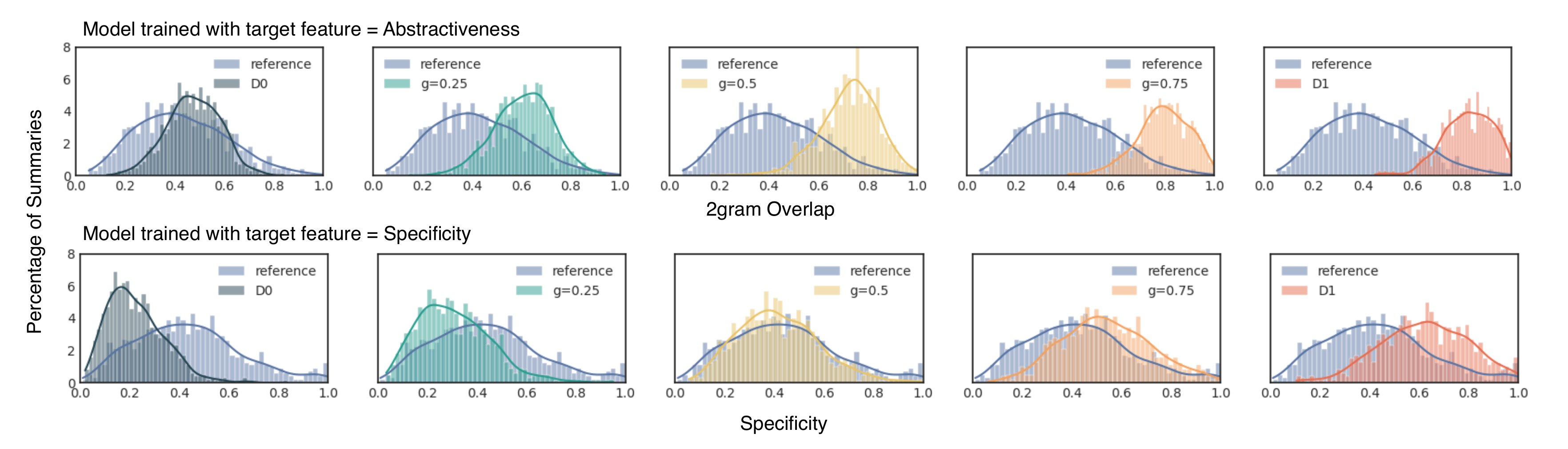} \vspace{-5mm}
    \caption{2gram overlap and specificity of \textsc{Cnn} outputs with different values of $g$ under extreme partitioning. The top graphs are from the $f=$ abstractiveness and the bottom are from the $f=$ specificity model. For each, the leftmost graphs correspond to low-$f$ (D0) decoders; the contribution of high-$f$ (D1) increases as we move right. These graphs clearly show that target features can be reliably varied by varying gate probabilities.}
    \label{fig:cnn-guided}
\end{figure*} 

\paragraph{Can we use \modelname~to vary summary styles between these \textit{extremes}?} To study this, we generate 5 summaries for each input by varying the gate probabilities: $g=\{0, .25, .5, .75, 1\}$. We plot the 2-gram overlap of \textsc{Cnn} summaries for the 5 different gate values for the $f=$ abstractiveness model. Similarly, we plot specificity for the $f=$ specificity model at different gate levels (see Figure \ref{fig:cnn-guided}). Due to space constraints, graphs for \textsc{Newsroom} and \textsc{XSum} are in Appendix \ref{appendix:guided-setting}. 

For both stylistic features, we observe that the \modelname~model shows a gradual increase in average feature scores as the contribution of D1 (high-$f$ decoder) is increased, from 0 contribution in the leftmost graphs to 1 in the rightmost graphs. This shows that \modelname~can be used to reliably vary style along a target feature. The graphs also show that our model can sample summaries from a wider area in the generation space compared to baseline models (i.e. compare the 2-gram overlap in Figure~\ref{fig:lexical-unguided} with the diversity of overlap in Figure~\ref{fig:cnn-guided}).

\paragraph{Can we \textit{mix} decoders of any two separately trained \modelname~models?} This further tests the flexibility of our models. Here, we run experiments that combine \modelname~decoders exhibiting extreme styles along orthogonal features of abstractiveness and specificity (from Section \ref{sec:experiments-guided}), but trained on the same dataset. Choice of such orthogonal styles aids our evaluation by providing a desiderata for generated summaries; if we combine the \textit{highly} extractive and \textit{highly} specific decoders from separate models, we want \modelname~to output summaries that follow both these properties. 

\begin{figure}[t]
\centering
    \includegraphics[trim=10mm 85mm 10mm 95mm,scale=0.23, clip]{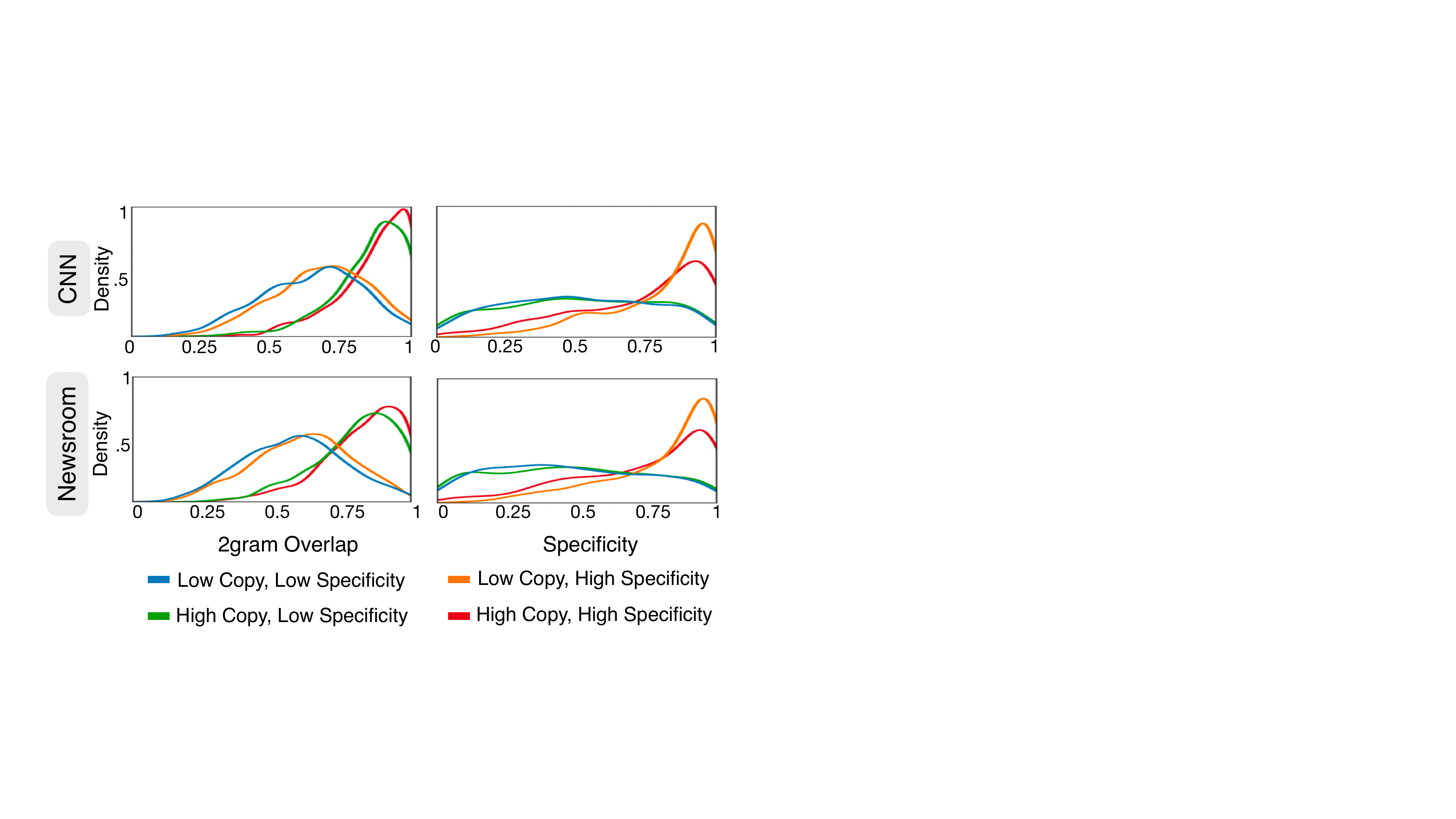} 
    \caption{2-gram overlap and specificity of \textsc{Cnn} and \textsc{Newsroom} summaries generated using combinations of $f=$ specificity and $f=$ abstractiveness decoders.}
    \label{fig:mult-attribute}
\end{figure} 

We conduct this experiment for \textsc{Cnn} and \textsc{Newsroom} datasets (\textsc{XSum} is omitted due to low separation along abstractiveness). We target the following pairs, setting gate probability $g = 0.5$: (1) high copy, low specificity, (2) low copy, low specificity, (3) low copy, high specificity and (4) high copy, high specificity. The marginal distribution of each feature for all four combinations is plotted in Figure \ref{fig:mult-attribute}; the left graphs plot 2gram overlap and the right graphs plot specificity. They show that the \modelname~summaries generated using a high specificity decoder in the mixture generates more specific  summaries on average compared to those using the low specificity decoders. Similar trends are observed for abstractiveness. These results expose potential new use cases of \modelname~models, including multi-feature control. We leave further exploration of this capability for future work.

\section{Human Evaluation} 
\label{sec:human-eval}
Following prior work \citep{hashimoto2019unifying}, we conduct human evaluation to measure the quality of generated summaries. For $50$ randomly sampled input articles from each dataset, we present MTurk workers with 5 different generated summaries: baseline model summary, D0 and D1 summaries of the $f=$ abstractiveness and $f=$ specificity models. The workers were asked to rate each summary along 4 dimensions: relevance, coherence, grammatically and factuality. For the first $3$, we ask for a rating on the 5-point Likert scale. Following \citet{goyal-durrett-2021-annotating}, we seek binary labels (factual (1) or non-factual (0)) for factuality annotation. More details and task interface are in Appendix \ref{appendix:human-eval}. We report the average score of all three annotations in Table \ref{tab:human-eval-guided}. Across all metrics, we see that the humans score summaries generated by the \modelname~models higher than the baseline models. Human annotation results corresponding to the summaries in Table \ref{table:experiments-unguided} are in Appendix \ref{appendix:human-eval}.

\begin{table}[t]
    \centering
    \small
    \begin{tabular}{c|c|cc}
    \toprule
        Data & Model & $f=$ Abs. &  $f=$ Spec. \\ \midrule
         \multirow{3}{*}{\textsc{Cnn}} & BS &  \multicolumn{2}{c}{4.3/4.4/4.2/.83}\\
         & HS D0 & \textbf{4.4/4.5/4.3/.93} & \textbf{4.4}/4.3/4.2/.85 \\
         & HS D1 & 4.3\textbf{/4.5/4.3}/.89 & \textbf{4.4}/4.3/4.1/.87 \\ \midrule
         \multirow{3}{*}{\textsc{NRoom}} & BS & \multicolumn{2}{c}{4.2/4.3/4.0/.85}\\
         & HS D0 & \textbf{4.3/4.4/4.1/.9} & 4.1/4.2/3.5/.80 \\
         & HS D1 & 4.2/4.2/4.0/\textbf{.9} & 4.2\textbf{/4.4/4.1/}.85\\ \midrule
         \multirow{3}{*}{\textsc{XSum}}& BS & \multicolumn{2}{c}{4.3/4.4\textbf{/4.2/}0.85}\\
         & HS D0 & 4.2/4.3/4.1/.89 & 4.3/4.4/4.1/.81\\
         & HS D1 & 4.3\textbf{/4.5/}4.0/.87 & \textbf{4.4/}4.4\textbf{/4.2/.89}\\
    \bottomrule
    \end{tabular}
    \caption{Human-annotated \textbf{Relevance/Coherence/ Grammaticality/Factuality} scores for $f=$ abstractiveness and  $f=$ specificity \modelname~ models. We report results for both decoders (D0 and D1) and compare against the baseline \textsc{Bart} model.}
    \label{tab:human-eval-guided}
\end{table}

\section{Related Work}
\label{sec:related-work}
Prior work on style control in summarization focuses on features like length \citep{fan2018controllable, song2021new}, abstractiveness \citep{song2020controlling}, etc. It has also been studied for other generation tasks such as paraphrasing and story generation \citep{wang2017steering, shen2017style, huang2019hierarchically}. These methods are over-specialized for the target style and cannot be easily generalized to more features. Recently, GeDi \citep{krause-etal-2021-gedi-generative} proposed using small LMs as generative discriminators for specific attributes (e.g. toxicity) to guide the generation of larger models. Similar class-conditional language models approaches (CC-LMs) have been previously proposed \citep{keskar2019ctrl, ficler2017controlling} to fine-tune models on specific attributes. Contrary to these, \modelname~models can disentangle styles within the task-specific datasets without explicit style annotations, as well as cover the generation space between two `extreme' styles.

Diverse generation has more widely been studied for other generation tasks, including decoding modifications   \citep{vijayakumar2018diverse, kumar2019submodular}, enforcing syntactic diversity \citep{goyal2020neural}, or through uninterpretable latent codes \citep{park2019paraphrase, shao2019long}. In this work, we study diversity in style that naturally emerges under standard training and decoding. 

\section{Conclusion}
We propose a new summarization architecture \modelname~containing multiple decoders in a mixture-of-experts. Our model automatically separates distinct summary styles, e.g. high or low abstractiveness, different levels of specificity, etc., across different decoders under the standard training regimen. We show that the proposed model is highly flexible; during inference, we can sample from either individual decoders or their mixtures to vary summary features.

\section{Limitations}
In this paper, we propose a simple modification to existing summarization architectures to disentangle style features. Although this modification is not language-dependant, all our experimentation and analysis is performed only on English language summarization datasets. Furthermore, we only study newswire summaries due to their popularity in summarization research. Therefore, this paper does not provide insights into what style diversity exists in non-English and non-newswire datasets, or whether our findings generalize to these other datasets. 

Next, we study style partitioning along a limited number of style dimensions, both due to computational constraints, as well as space constraints in the paper. Due to similar computational constraints, we run all our experiments using the \textsc{Bart} model as a case study. While we strongly believe that our conclusions are generalizable to other pre-trained models like \textsc{Pegasus}, we do not show explicit evidence for this. Note that multiple prior works in summarization have discussed that both \textsc{Bart} and \textsc{Pegasus} exhibit similar high-level trends across various summarization behaviors \cite{xu2020understanding, goodwin2020flight}.

\section*{Acknowledgments}
We thank Greg Durrett, Jessy Li and Jiacheng Xu for reviewing an earlier version of this paper and providing valuable feedback. Thanks as well to the Amazon Mechanical Turk workers for participating in the human annotation study and the anonymous reviewers for their helpful comments.

\bibliography{anthology,custom}
\bibliographystyle{acl_natbib}

\appendix

\section{Training Details}
\label{appendix-trainingdetails}
\begin{table}[h]
    \vspace{-1em}
    \centering
    \small
    \begin{tabular}{c|ccc}
        \toprule
        Dataset & Training & Dev & Test \\ \midrule
         \textsc{Cnn} & 90266 & 1220 & 1093 \\
        \textsc{Newsroom} &  329494 & 35977 & 36100 \\
         \textsc{XSum} & 204045 & 11332 & 11334 \\
         \bottomrule
    \end{tabular}
    \caption{Dataset statistics}
    \label{table:dataset-statistics}
\end{table}

\begin{table*}[t]
    \small
    \centering
    \begin{tabular}{r|l||r|l}
        \toprule
        \multicolumn{2}{c||}{\textbf{For training}} & \multicolumn{2}{c}{\textbf{For Inference}} \\ \midrule
        Implementation & Huggingface \citep{wolf2020transformers} & \multicolumn{2}{c}{\cellcolor{Gray} \textsc{Cnn \& Newsroom}} \\
        Infrastructure & 40 GB NVIDIA A100 GPU & Num beams & 5\\
        Optimizer & Adam & Length Penalty & 2\\
        Optimizer Params & $\beta= (0.9, 0.999), \epsilon = 10^{-8}$ & No repetition size & 3-grams \\
        Learning Rate Decay & Linear & Min-Length & 12 \\
        Learning rate & 1e-5** & Max Length & 200 \\
        Weight Decay & 0 &  \multicolumn{2}{c}{\cellcolor{Gray} \textsc{Xsum}} \\
        Maximum Gradient Norm & 1 & Num beams & 6 \\
        Batch size & 64 & Length Penalty & 1   \\
        Epochs & 3 & No repetition size & 3-grams\\
        Max Input Length & 1024 (512 for \textsc{Newsroom}) & Min Length & 12  \\
        Max Output Length & 128 & Max Length & 60 \\
        \bottomrule
    \end{tabular}
    \caption{Hyperparameters used for fine-tuning and decoding the \textsc{Bart}-based summarization models. (**For $f=$ specificity models in Section \ref{sec:experiments-guided}, we set learning rate to 2e-5)}
    \label{table:hyperparameters}
\end{table*}

\begin{table*}[t]
    \small
    \centering
    \begin{tabular}{c|c|cc|cc|cc|cc}
        \toprule
        Dataset & m & \multicolumn{2}{c|}{\textsc{Rouge}} & \multicolumn{2}{c|}{Overlap} & \multicolumn{2}{c|}{Specificity} & \multicolumn{2}{c}{Length} \\ \midrule
        & & D0 & D1 & D0 & D1 & D0 & D1 & D0 & D1 \\ 
        \midrule
         \multirow{2}{*}{\textsc{Cnn}} & 6 & 33.21/13.3/30.21 & 34.26/13.30/31.21 & \cellcolor{Gray}.79 & \cellcolor{Gray}.63 & .42 & .43 & \cellcolor{Gray}44.9 & \cellcolor{Gray}54.5\\
         & 10 & 32.04/12.37/29.13 & 35.20/14.11/32.19 & \cellcolor{Gray}.80 & \cellcolor{Gray}.68 & \cellcolor{Gray}.38 & \cellcolor{Gray}.45 & \cellcolor{Gray}53.8 & \cellcolor{Gray}45.9 \\ \midrule
         \multirow{2}{*}{\textsc{Newsroom}} & 6 & 32.32/16.17/27.50 & 34.92/17.05/29.55 & \cellcolor{Gray}.82 & \cellcolor{Gray}.61 & .60 & .60 &  \cellcolor{Gray}39.5 & \cellcolor{Gray}30.0 \\
         & 10 & 33.14/16.56/28.16 & 34.73/17.10/29.37 & \cellcolor{Gray}.79 & \cellcolor{Gray}.64 & \cellcolor{Gray}.57 & \cellcolor{Gray}.64 & 33.9 & 34.6\\ \midrule
         \multirow{2}{*}{\textsc{XSum}} & 6 & 42.20/18.70/33.60 & 42.30/18.70/33.90 & .22 & .23 & \cellcolor{Gray}.66  & \cellcolor{Gray}.53 & 20.2 & 19.8 \\
         & 10 & 42.56/19.14/34.10 & 42.83/19.15/34.24 & .24 & .23 & \cellcolor{Gray}.64 & \cellcolor{Gray}.56 & 19.0 & 20.5 \\
         \bottomrule
    \end{tabular}
    \caption{Effect of varying the number of shared layers between the 2 decoders of \modelname. Results show that the choice of $m$ does not substantially alter our analysis.}
    \label{table:effect-different-m}
\end{table*}

We evaluate our models on three datasets: \textsc{Cnn}, \textsc{Newsroom} and \textsc{XSum}. Training, development and test dataset sizes for these are listed in Table \ref{table:dataset-statistics}. Note that our experiments (both training and evaluation) are performed on the \textit{mixed} subset of the \textsc{Newsroom} dataset. All results and analysis in the paper is reported on the test data.

Table \ref{table:hyperparameters} outlines the hyperparameters used for training and inference. For all our experiments, we use \textsc{Bart-Large} as the pre-trained initialization. During inference for \modelname, we incorporate top-k and top-p sampling using values 30 and 0.5 respectively. For top-k decoding using baseline \textsc{Bart} model in Table \ref{tab:diversity-eval-rouge}, we set $k=30$. Diverse beam search is run using 2 beam groups and diversity penalty $0.5$.

\section{Effect of different number of shared layers}
\label{appendix:varying-m}
In order to restrict the number of extra parameters introduced in \modelname, we enforced parameter sharing between the $m$  lower layers of the decoders. We performed our all experiments in Section \ref{sec:experiments} and \ref{sec:experiments-guided} by setting $m=8$. Here, we investigate if the choice of $m$ effects either the partitioning of stylistic features between decoders, or the extent of the observed difference between two decoders along any axis such as abstractiveness, specificity, etc. Experiments are additionally performed using the 2-decoder version of \modelname for $m=6, 10$ for all 3 datasets. For simpler analysis, we only report on a subset of the metrics: \textsc{Rouge} scores (quality), 2 gram overlap (abstractiveness), specificity and absolute length between the summaries generated using individual decoders. 

Table \ref{table:effect-different-m} outlines the results. Compared to the \modelname~model variants with $m=8$, we notice small differences in style partitioning as well as the absolute difference in style scores between decoders D0 and D1. Most notably, the \textsc{Cnn} and \textsc{Newsroom} model with $6$ shared parameters does not learn to partition across the specificity metric whereas the \textsc{Newsroom} model with $m=6$ does learn to partition along length. These observations are different that those seen for $m=8,10$. However, in general, we observe that across all datasets, \modelname~decoders behave quite similarly in terms of which features are partitioned, irrespective of the number of shared layers $m$. This demonstrates that the proposed model architecture is useful for generating diverse summary options, even in cases where a smaller number of extra parameters are allowed.

\section{Human Evaluation}
\label{appendix:human-eval}
\begin{figure*}[h]
\centering
\includegraphics[trim=130mm 40mm 0mm 25mm,scale=0.75, clip]{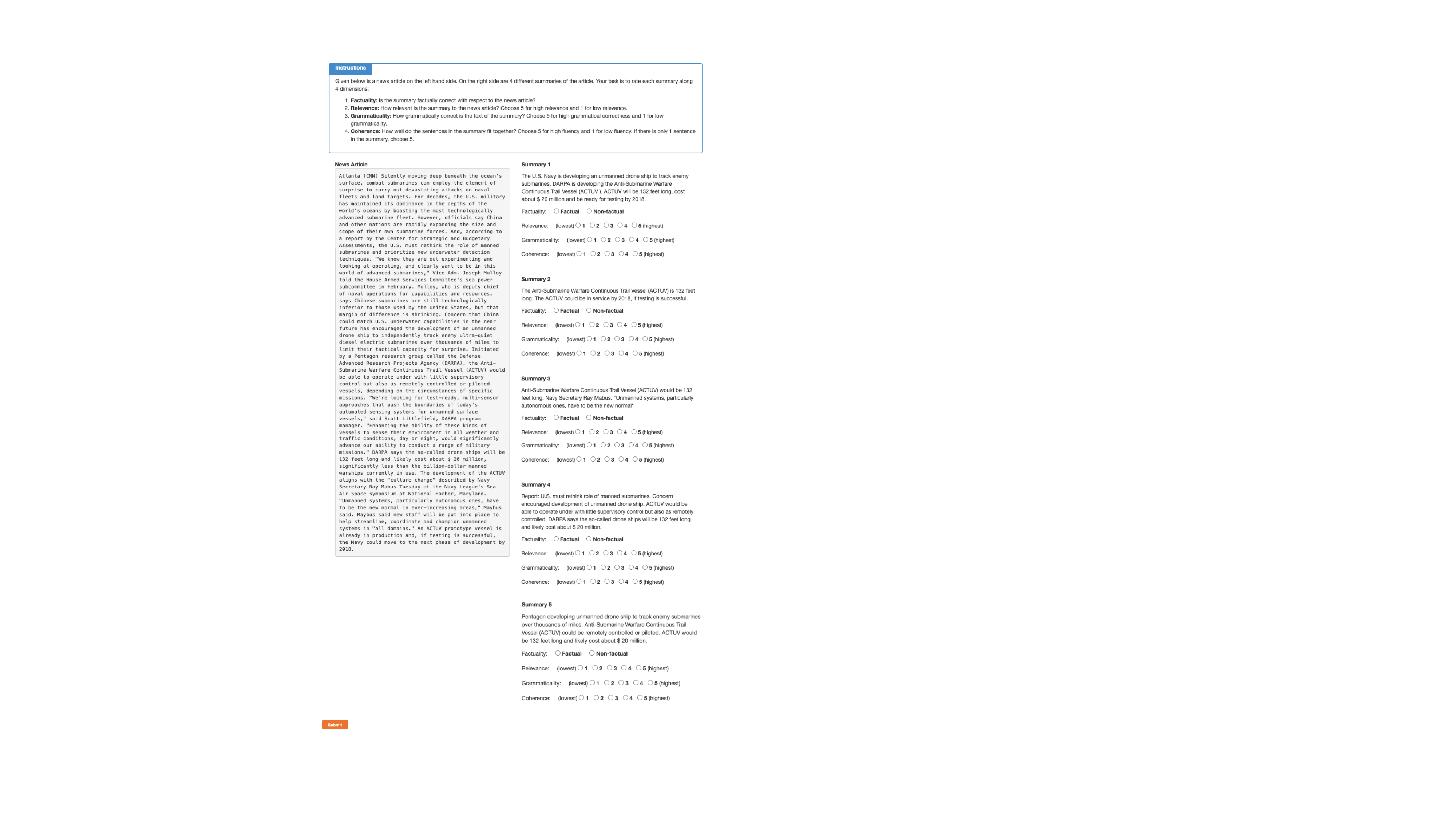}
\caption{Interface of the Mechanical Turk Task}
\label{fig:mturk-ui}
\end{figure*}

In section \ref{sec:experiments-guided}, we reported human evaluation study results under extreme partitioning. Here, we expand on the details of the Mechanical Turk task. Figure \ref{fig:mturk-ui} shows task interface. For each source article, we asked $3$ workers to evaluate $5$ different model-generated summaries. For the extreme partitioning setting, these 5 summaries were obtained from (1) Baseline model, (2, 3) D0 and D1 decoders of the $f=$ abstractiveness model, and (4,5) D0 and D1 of the $f=$ abstractiveness model. For each article-summary pair, workers were asked to rate the summaries across 4 metrics: relevance, coherence, grammaticality, and factuality. We follow prior work \citep{karpinska2021perils} and seek annotation for the first $3$ on a 5-point Likert scale, with $5$ corresponding to highest quality. For factuality, we ask for a binary annotation: $1$ for factuality and $0$ for non-factual summaries. We report the average scores of the 3 annotators across all 50 articles, for each dataset. 

\begin{table}[h]
    \centering
    \small
    \begin{tabular}{cccc}
    \toprule
        ~ & \textsc{Cnn} & \textsc{Newsroom} & \textsc{XSum} \\ \midrule
        D0 & 4.3/\textbf{4.4/4.2}/.86 & \textbf{4.4/}4.4/4.2/\textbf{.92} & \textbf{4.3/}4.3/4.2\textbf{/.81} \\ 
        D1 & 4.3/4.3/4.0/\textbf{.89} & 4.2/4.4/4.1/.91 & 4.1/4.4/4.2/.81 \\
        Mix & \textbf{4.4}/4.3/4.2/.87 & \textbf{4.4/4.5/4.3}/.9 & 4.2/\textbf{4.5/4.3}/.8 \\ \midrule
        BS & \textbf{4.4/4.4/4.2}/.88 & 4.3/4.4/4.2/.9 & \textbf{4.3}/4.4/4.2/.77 \\
    \bottomrule
    \end{tabular} 
    \caption{Comparison of human-annotated \textbf{Relevance/Coherence/Grammaticality/Factuality} scores of \modelname~models (using individual decoders D0 and D1, and their mixture) and baseline \textsc{Bart} (BS).} 
    \label{tab:human-eval-unguided}
\end{table}

Next, we conducted an analogous study for our original training setting, corresponding to the standard training regimen. For this, we asked workers to rate the quality of $4$ different summaries per article (1) baseline model, (2, 3) D0 and D1 of \modelname~model, and (4) Mix strategy of \modelname~model. Again, we ask ratings for $50$ randomly sampled articles (note that these articles are different from the ones annotated in the baseline setting, and therefore, baseline model results may differ). Table \ref{tab:human-eval-unguided} outlines the results. The results show that the \modelname~model performs on par with the baseline model along all quality dimensions measured, even outperforming it in terms of factuality for both \textsc{Newsroom} and \textsc{Xsum}. This agrees with our results from Table \ref{table:experiments-unguided} which similarly shows that both the baseline and \modelname~model summaries have similar quality.

\section{Extreme Partitioning - Additional Results}
\label{appendix:guided-setting}
In Section \ref{sec:experiments-guided}, we reported the style scores of the different models under our \textit{extreme} partitioning scenario. Table \ref{table:experiments-guided-small} outlined a brief summary of results for models trained on the three datasets. Here, we provide the entire set of results, see Table \ref{table:experiments-guided}. In addition to the metrics reported in the main paper, we include \textsc{Rouge} scores of individual decoders D0 and D1 for both $f \in$ \{ abstractiveness, specificity\}  models. Moreover, other style metrics (in addition to the target $f$ of each model) are also included for each model and dataset pair (2-gram overlap, specificity and length). Table \ref{table:experiments-guided} outlines the results. In general, we observe that \modelname~models are able to enforce diverse generation along the target feature $f$, while limiting the stylistic variance along other features between D0 and D1.
Figure \ref{fig:examples-specificity} includes examples of low- and high-specificity summaries generated using the $f=$ specificity model.

\begin{table}[t]
    \small
    \centering
    \begin{tabular}{c|c|c|ccc}
    \toprule
        $f$& Dec. & Quality & \multicolumn{3}{|c}{Summary Styles} \\ \midrule
        & & \textsc{Rouge} & Ov. & Sp. & Len \\ \midrule
        \rowcolor{Gray}
        \multicolumn{6}{l}{\textsc{Cnn}} \\  
        \multirow{2}[1]{*}{Abs.} & D0 & 35.00/12.93/31.84  & .48$^\dagger$ & .42 & 48.8 \\
         & D1 &34.66/14.45/31.78 & .82$^\dagger$ & .42 & 46.2 \\ \midrule
        \multirow{2}[1]{*}{Spec.} & D0 & 33.64/12.74/30.70 & .72 & .22$^\dagger$ & 48.9\\
        & D1 & 34.40/13.35/31.18 & .69 & .62$^\dagger$ & 49.7  \\ \midrule
        \rowcolor{Gray}
        \multicolumn{6}{l}{\textsc{Newsroom}} \\ 
        \multirow{2}[1]{*}{Abs.} & D0 & 32.56/13.98/26.68 & .44$^\dagger$ & .65 & 35.8 \\
         & D1 & 35.04/18.53/30.17 & .85$^\dagger$ & .59 & 33.9  \\ \midrule
        \multirow{2}[1]{*}{Spec.} & D0 & 31.62/14.80/27.11 & .67 & .36$^\dagger$ & 27.0 \\
        & D1 & 34.20/17.26/28.74 & .73 & .81$^\dagger$ & 38.4 \\ \midrule
        \rowcolor{Gray}
       \multicolumn{6}{l}{\textsc{XSum}} \\ 
        \multirow{2}[1]{*}{Abs.} & D0 & 42.45/19.00/34.35 & .16$^\dagger$ & .58 & 19.2 \\
         & D1 & 43.52/19.79/35.05 & .29$^\dagger$ & .57 & 19.5 \\ \midrule
        \multirow{2}[1]{*}{Spec.} & D0 & 41.84/18.55/33.86 & .22 & .44$^\dagger$ & 18.2\\
        & D1 & 41.72/18.14/33.11 & .22 & .80$^\dagger$ & 21.8\\
    \bottomrule
    \end{tabular}
    \caption{Performance of \textit{extreme} partitioned \modelname~models. Compared to Table \ref{table:experiments-unguided}, we observe higher variation in style between D0 and D1 along the target dimension (indicated with $\dagger$)}
    \label{table:experiments-guided}
\end{table}

\begin{figure*}[t]
\centering
\subfloat[][\textsc{Newsroom}]{\includegraphics[trim=50mm 160mm 0mm 40mm,scale=0.28, clip]{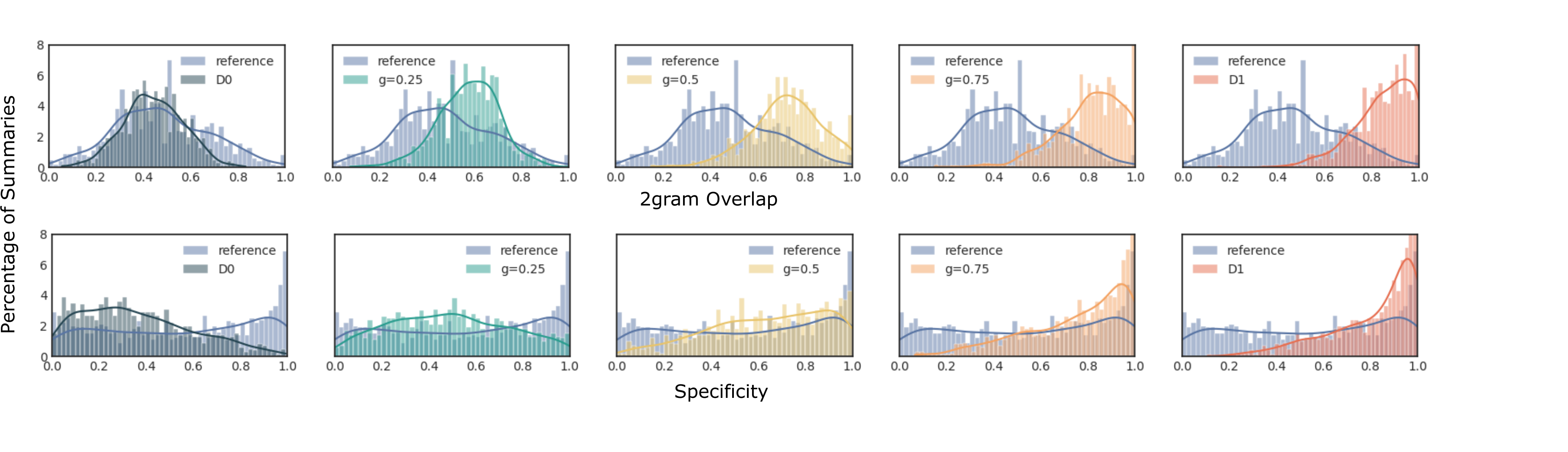}}\\ 
\subfloat[][\textsc{Xsum}]{\includegraphics[trim=45mm 160mm 0mm 40mm,scale=0.28, clip]{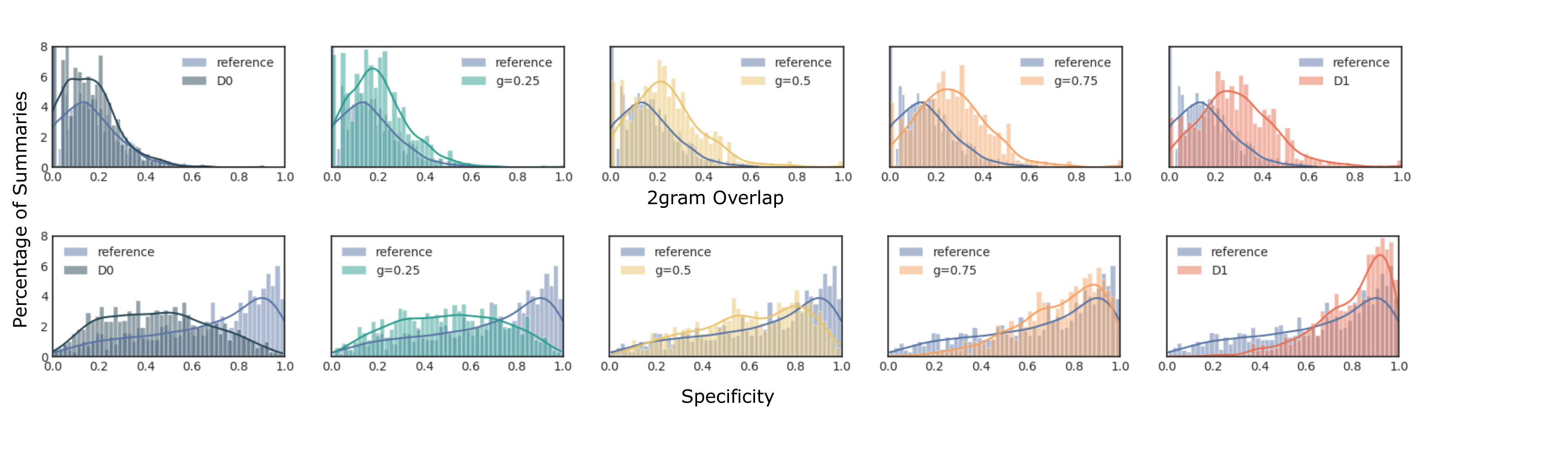}}
\caption{2gram overlap and specificity of \textsc{Newsroom} and \textsc{XSum} summaries generated using different values of $g$. The graphs show that properties like abstractiveness and specificity can be varied by sampling from a mixture of the 2 decoders corresponding to the chosen style.}
\label{fig:newsroom-xsum-guided}
\end{figure*}

Finally, in Figure \ref{fig:newsroom-xsum-guided}, we include graphs that show the distributions of 2 gram overlap and specificity for the $f=$ abstractiveness (top row) and $f=$ specificity (bottom row) models respectively, for datasets \textsc{Newsroom} and \textsc{XSum} models. The corresponding graphs for \textsc{Cnn} are included in the main body of the paper (section \ref{sec:experiments-guided}).

\begin{figure*}[t]
\centering
\includegraphics[trim=40mm 145mm 0mm 70mm,scale=0.28, clip]{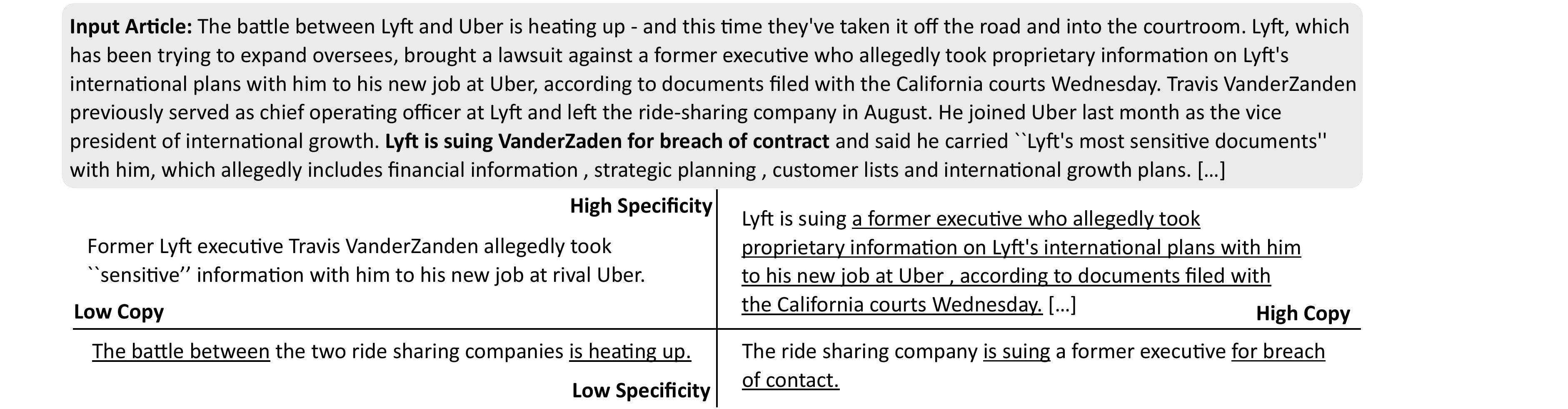}
\caption{Examples of summaries generated by combining \modelname~decoders from different models and corresponding to different extreme styles.}
\label{fig:mult-attribute-ex}
\end{figure*}

\section{Combining multi-feature decoders}
\label{appendix:ex-multi-feature}
Figure \ref{fig:newsroom-xsum-guided} shows an example of summaries generated using a combination of \textit{extreme} decoders corresponding to orthogonal features for the \textsc{Newsroom} dataset. We 4 generate summaries by using a distinct combination of extractive/abstractive and general/specific decoders from different single-feature controlled models. The figure shows the input article and these generated summaries: we see that these summary follow the style specifications of the two decoders used to construct them. Interestingly, for the High Copy, Low specificity summary, we see that the model replaces \textit{Lyft} with \textit{ride-sharing company} and \textit{VanderSaden} with \textit{former executive} from an exact copied sentence from the input, to both follow high copy and low specificity targets as faithfully as possible. In general, we found summary generation including a low specificity decoder tougher to control (here, the Low copy, Low Specificity summary follows similar strategy to the High Copy, Low Specificity summary). This is also evidenced by specificity distributions in Figures \ref{fig:newsroom-xsum-guided} which show much higher variation for D0 (i.e. low specificity decoder) for the specificity controlled model. Similar trends are seen in Figure \ref{fig:mult-attribute}.

\end{document}